\documentclass[10pt,twocolumn,letterpaper]{article}

\usepackage[pagenumbers]{cvpr}      %

\usepackage{graphicx}
\usepackage{amsmath}
\usepackage{amssymb}
\usepackage{booktabs}
\usepackage{bm}
\usepackage{times}   %
\usepackage{cvpr} 
\usepackage{bm}
\usepackage{multirow}
\usepackage{makecell}
\usepackage{xcolor} 
\usepackage[normalem]{ulem}
\usepackage{balance}

\newcommand{\seql}[1]{{\textcolor{gray}{#1}}}

\newcommand{\fig}[1]{Figure~\ref{#1}}

\newcommand{\tbl}[1]{Table~\ref{#1}}
\newcommand{\xpar}[1]{\noindent{\textbf{#1}}}

\def\eg{\emph{e.g.}} 
\def\ie{\emph{i.e.}}

\def\etal{\emph{et al.}}

\DeclareMathOperator*{\argmax}{arg\,max}

\newcommand{\supp}{the supplementary material\xspace}

\newcommand{\privd}[1]{\textcolor{blue}{#1}}

\usepackage{pifont}%
\newcommand{\cmark}{\textcolor{green}{\ding{51}}}%
\newcommand{\xmark}{\textcolor{red}{\ding{55}}}%

\usepackage[pagebackref,breaklinks,colorlinks]{hyperref}

\usepackage[capitalize]{cleveref}
\crefname{section}{Sec.}{Secs.}
\Crefname{section}{Section}{Sections}
\Crefname{table}{Table}{Tables}
\crefname{table}{Tab.}{Tabs.}

\begin{document}

\title{Sub-word Level Lip Reading With Visual Attention}

\author{K R Prajwal   \quad Triantafyllos Afouras    \quad {Andrew Zisserman}  \\     Visual Geometry Group, University of Oxford \\       \texttt{\{prajwal,afourast,az\}@robots.ox.ac.uk} \\ } 

\maketitle

\begin{abstract}
The goal of this paper is to learn strong lip reading models that can recognise speech in silent videos. Most prior works deal with the open-set visual speech recognition problem by adapting existing automatic speech recognition techniques on top of trivially pooled visual features. Instead, in this paper we focus on the unique challenges encountered in lip reading and propose tailored solutions. To this end, we make the following contributions: (1) we propose an attention-based pooling mechanism to aggregate visual speech representations; (2) we use sub-word units for lip reading for the first time and show that this allows us to better model the ambiguities of the task; (3) we propose a model for Visual Speech Detection (VSD), trained on top of the lip reading network. Following the above, we obtain state-of-the-art results on the challenging LRS2 and LRS3 benchmarks when training on public datasets, and even surpass models trained on large-scale industrial datasets by using an order of magnitude less data. Our best model achieves 22.6\% word error rate on the LRS2 dataset, a performance unprecedented for lip reading models, significantly reducing the performance gap between lip reading and automatic speech recognition. Moreover, on the AVA-ActiveSpeaker benchmark, our VSD model surpasses all visual-only baselines and even outperforms several recent audio-visual methods.
\end{abstract}

\section{Introduction}

Lip reading, or visual speech recognition, is the task of recognising speech from silent video.
It has many practical applications 
which include improving speech recognition in noisy environments, 
enabling silent dictation,
or dubbing and transcribing archival silent films~\cite{Jha2019}. It also has important medical applications, such as helping speech impaired
individuals, e.g.\ people suffering from Lou Gehrig's disease
speak~\cite{Shillingford18}, or enabling people with aphonia (loss of voice) to communicate just by using lip movements.

Lip reading and audio-based automatic speech recognition (ASR)
both have the common goal of transcribing speech, however they differ 
regarding the input:
while in ASR the input signal is an audio waveform, in essence a one-dimensional time-series,
lip reading has to deal with high-dimensional video inputs that
have both temporal and spatial complexity.
This makes training large end-to-end models harder
due to GPU memory and computation constraints.
Furthermore, understanding speech from visual information alone is challenging due to the inherent
ambiguities present in the visual stream, i.e.\ the existence of homophemes where
different characters that are visually indistinguishable (e.g. `pa', `ba' and `ma').
That lip reading is a much harder task is also supported by
the fact that although humans can understand speech reasonably well even in the presence of noise
and across a variety of accents, they perform relatively poorly on lip reading~\cite{Chung16,Assael16}.

Designing a lip reading model requires both a visual component -- mouth movements need to be identified --
as well as a temporal sequence modelling component, which typically involves learning a language model that can resolve ambiguities in individual lip shapes.
Recent developments in deep learning models and the availability of large-scale annotated datasets 
has led to breakthroughs surpassing human 
performance~\cite{Chung16}.
However, most of these works have taken the approach of adapting
techniques used for ASR and machine translation, without catering
to the particularities of the vision problem.

The conjecture in this paper is that the performance of lip reading, in terms of both accuracy and data efficiency, can be improved if the model is designed from the start taking account of the peculiarities of the visual, rather than the audio domain. To this end, we consider both the visual encoding and the text tokenisation. 

\noindent{\textbf{Visual encoding.}} 
Our first contribution is the design of a novel visual backbone for lip reading.
The spatio-temporal complexity in lip reading requires
dealing with problems such as tracking the mouth in moving talking heads.
This is usually achieved with complicated pre-processing pipelines based on facial landmarks.
However, those are sub-optimal in many cases. For example, landmarks don't work well in profile views~\cite{Johnston18landmarks}.
Moreover, it is unclear what is the optimal region-of-interest for lip reading: %
it has
been shown that besides the lips, other parts of the face, e.g. the cheeks, may also contain useful discriminative information~\cite{zhang2020read}. Also, this region-of-interest can vary drastically in terms of scale, aspect ratio across identities and utterances.
Thus, in this work, we propose an end-to-end trainable attention-based pooling mechanism that learns to track and
aggregate the lip movement representations, resulting in a significant performance boost.

\noindent{\textbf{Text tokenisation.}}
Lip reading methods most commonly output character-level tokens.
This output representation however is sub-optimal as characters 
are sometimes more fine-grained than the input, with multiple characters corresponding to a single video frame.
Furthermore, characters do not encode any prior knowledge about the language, which leads to higher dependency on the decoder's language modeling capacity that must also `learn to read'.
In this work we instead use sub-word tokens (word-pieces) which 
not only match with multiple adjacent frames but are also semantically meaningful for learning a language easily. 
Word-pieces result in much shorter (than character) output sequences which greatly 
reduces the run-time and memory requirements. They 
also provide a language prior, reducing the language modelling burden of the model.
We experimentally compare character and word-piece tokenization to justify this choice.

\xpar{Visual Speech Detection.}
One issue with performing lip reading inference on real-world silent videos is that, since there is no audio track, there is no automated procedure for 
cropping out the clips where the person is speaking. 
ASR models use Voice Activity Detection (VAD) as a key pre-processing step, but this is clearly not applicable for silent videos. Here, the parts of a video containing speech have to be determined using the video input alone; in other words, by performing Visual Speech Detection (VSD).
This can be very useful \eg~for running inference on silent movies.
Among other findings in this work, we show that it is possible to train a strong VSD model on top of our pre-trained lip reading encoder.

\xpar{Other downstream tasks.}
Besides improving performance on the sentence-level lip reading task itself,
obtaining improved lip movement representations can have broader impact, as those are
often used for other related downstream tasks -- e.g.\
sound source separation~\cite{ephrat2018looking},
visual keyword spotting~\cite{Momeni20},
and visual language identification~\cite{Afouras20c}.

In summary, we make the following three contributions: 
(i) a visual backbone architecture using attention based pooling on the spatial feature map; (ii) the use of sub-word units, rather than characters for the language tokens; and 
(iii) a strong Visual Speech Detection model, directly trained on top of the lip reading encoder.

In the experiments we show the benefits of (i) and (ii) on improving lip reading performance, and we also introduce a two stage training protocol that simplifies the curriculum used in prior works.
As will be seen, with these design choices and training methodology, the performance of our best models exceeds prior work on standard evaluation benchmarks, and even outperforms proprietary models that use an order of magnitude more data for training. Similarly, we show the benefit of (i) and the lip reading encoder to our visual speech detection model that is far superior to previous methods on a standard evaluation benchmark.

We discuss potential ethical concerns and limitations of our work in \supp. 
Upon publication, we will make the code and pre-trained models public.

\section{Related Work}

We present an overview of prior work on lip reading,
including a discussion of how these methods select and track
the visual regions of interest, as well as the output tokenizations they use,
followed by a brief overview of the use of attention for visual feature aggregation in other domains.

\xpar{Lip reading.}
Early works on lip reading relied on hand-crafted pipelines and statistical models for visual feature extraction and temporal modelling~\cite{potamianos03,gowdy2004dbn,papandreou09,livescu2007articulatory,Ong11}; an extensive review of those methods is presented in~\cite{Zhou14}.
The advent of deep learning and the availability of large-scale lip reading datasets such as LRS2~\cite{Chung17} and LRS3~\cite{Afouras18d}, rejuvenated this area. Progress was initially on word-level
recognition~\cite{Chung16,Stafylakis17}, and then moved onto sentence-level recognition by adapting
models developed for ASR using LSTM sequence-to-sequence~\cite{Chung17} or
CTC~\cite{Assael16,Shillingford18} approaches.~\cite{petridis2018audio} take a hybrid approach, training an LSTM-based sequence-to-sequence
model with an auxiliary CTC loss.
One trend in recent work is moving to Transformer-based architectures~\cite{Afouras19}, or variants using convolution blocks~\cite{zhang2019spatio}, and hybrid architectures
like a Conformer~\cite{gulati2020conformer}. Another trend is to investigate the benefits of
training with larger datasets, either directly 
by training on proprietary data that is orders of magnitude larger than any public dataset~\cite{makino2019recurrent}, or indirectly by distilling ASR 
models into lip reading ones~\cite{Afouras20,Jianwei20tdnn,li2019improving}.
For visual feature extraction and short-term dynamic modelling, most modern pipelines rely on spatio-temporal CNNs consisting of multiple 3D convolutional layers~\cite{Assael16,Shillingford18}, or more lightweight alternatives that comprise a single 3D convolutional layer followed by 2D ones~\cite{Chung16,Stafylakis17,Afouras19} applied frame-wise.

\xpar{Mouth ROI selection, registration and tracking.}
A thorough investigation on facial region of interest (ROI) selection for lip reading is provided by~\cite{zhang2020read}. 
The videos included in datasets like LRS2 and LRS3 are commonly pre-processed 
with a face detection and tracking pipeline which outputs clips roughly centered around the
speaker's face.
Many previous works use a central crop on the provided videos
as input to the feature extractors~\cite{Stafylakis17, Afouras19, ma21conformer}.
More elaborate pipelines use facial landmarks to register the face to a canonical view
and/or only extract the crops of the mouth area~\cite{Assael16, Koumparoulis2017, Shillingford18, Petridis18,
makino2019recurrent, zhang2019spatio}.
~\cite{zhang2020read} propose inputting a large part of the face,
combined with Cutout\cite{devries2017improved} to encourage the
model to also use the extra-oral face regions.
After selecting which input region to extract the low-level CNN features from,
all above works apply Global Average Pooling (GAP) on the extracted visual features map;
this obtains a compact representation, but discards spatial information.
Recent works~\cite{zhang2019spatio} have shown that 
replacing GAP with a spatio-temporal fusion module improves performance.

\xpar{Text tokenization.}
Most prior works on lip reading output character-level
predictions~\cite{Chung17, Chung18a, Afouras19, zhang2019spatio, petridis2018audio, ma21conformer, makino2019recurrent}.
Those approaches usually use an external language model during inference to boost performance~\cite{Kannan17,Maas15}.
Instead~\cite{Shillingford18}
chose to output phoneme sequences, using phonetic dictionaries. 
This approach has the advantage of a more accurate mapping of lip-movements to sounds, 
but requires a complicated decoding pipeline involving a proprietary finite-state-transducer.~\cite{Koller15lip,Souheil20} use a hard-crafted heuristic to map words onto viseme sequences and vice versa, 
and use viseme tokens for representing the output and target text.  
In this work, we instead propose using sub-word level tokenisation, which greatly reduces the output
sequence length, thus accelerating both training and inference, and neatly encodes prior language
information that improves the overall performance.

\xpar{Visual feature aggregation with attention.}
Our work is also related to methods that use attention for improving visual representations of
images or videos. 
~\cite{Wang18nonlocal,jetley2018learn} use attention-weighted-averages of visual features as
building blocks for various classification and detection tasks, while 
OCNet~\cite{YuanW18} uses self-attention to model context between pixels for semantic segmentation.
A number of recent papers has replaced convolutions with Transformer~\cite{Vaswani2017} blocks in 
visual representation pipelines.
DETR~\cite{carion2020endtoend} and Efficient DETR~\cite{yao21efficientdetr} learn object detectors by
applying spatial transformers on top of CNN feature extractors. 
Similarly, the Visual Transformer~\cite{Wu20visual} tokenises low-level CNN features and then
processes them using a Transformer to model relationships between tokens.
ViT~\cite{dosovitskiy2021an} completely replaces CNNs in the visual pipeline with
Transformer layers applied on image patch sequences,
while the Timesformer~\cite{bertasius2021spacetime} has been suggested as a purely Transformer-based
solution for video representation learning. 

\xpar{Speech Detection.}
An important pre-processing stage in ASR pipelines is Voice Activity Detection (VAD),
which involves the detection of the presence of speech in audio~\cite{ramirez2007voice}.
The reliability of audio-based VAD systems deteriorates in the presence of noise or in cocktail party scenarios~\cite{Liu14vvad}.
In audio-visual pipelines, such as ones used for creating large-scale audio-visual speech datasets~\cite{Chung16,Afouras18d}, this step is commonly replaced by an Active Speaker Detection (ASD) stage which determines face tracks that match the speech. 
Audio-visual ASD models have been effectively trained 
either using direct supervision~\cite{Roth20ava,alcazar20,chung2019naver,leon2021maas,tao2021someone} or
in a self-supervised \cite{Chung16,Afouras20b} fashion by employing contrastive objectives. The visual counterpart to VAD is Visual Speech Detection (VAD), which operates only on the video input.
Early work on VSD (also termed Visual VAD or V-VAD) was based on handcrafted visual features
and statistical modelling using methods such as HMMs, GMMs
and PCA~\cite{Potamianos04,Liu04vvad,Siatras09vvad,Sodoyer06vvad,Sodoyer09vvad,Aubrey07vvad,Liu11vvad,Liu14vvad,Patrona16vvad}.
More recent works proposed methods based on
optical flow~\cite{Aubrey2010VisualVA} or a combination of CNN and LSTMs~\cite{sharma19vvad,guy2021learning}.
These methods are limited in having been trained or evaluated on constrained or non-public datasets. 
The train set of WildVVAD~\cite{guy2021learning}, a new annotated VSD dataset has been made public, however at the time
of submission its test set was not available,
we were therefore unable to use this dataset for benchmarking.

\section{Method}

\begin{figure*}
\centering                            
\includegraphics[width=\textwidth]{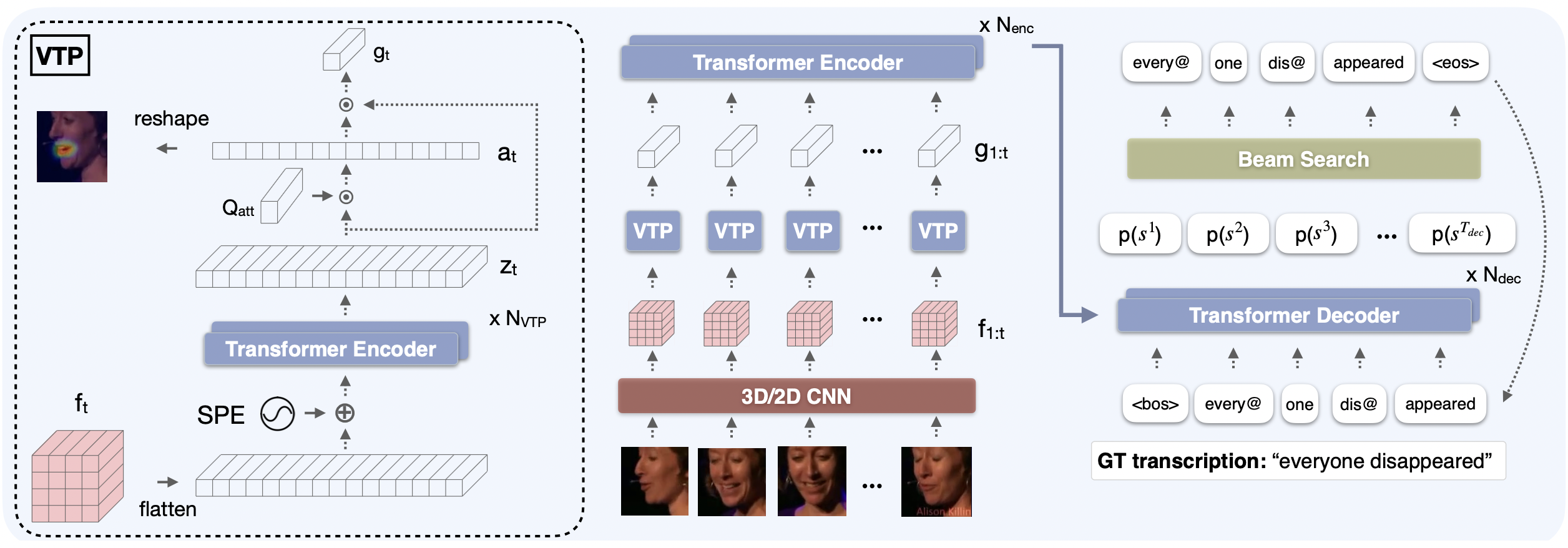} 
\caption{
  \textbf{Proposed lip reading architecture.} 
 {\em Left:} The input video frames are passed through a spatio-temporal CNN to extract low-level visual
  features $\bm{f}$. The feature map corresponding to every input frame is then separately processed by a
  Visual Transformer Pooling module (VTP).
  The VTP block adds spatial positional encodings (SPE)
  to the input features and passes the result through a Transformer encoder to produce a self-attended feature map $\bm{z}_t$.
  A query vector $\bm{Q_{att}}$ is used to compute an attention mask 
  which is in turn used to obtain a spatially weighted average of $\bm{z}_t$.
  This produces a compact visual representation of the lip appearance and movement around each input video frame.
  Concatenating the frame-wise features forms a temporal feature sequence $\bm{g}$.  
  This is passed as input to an encoder-decoder Transformer ({\em right})
  that auto-regressively predicts sub-word probabilities for one token at a time. 
  An output sentence is eventually inferred from these distributions using a beam search.
}
\label{fig:method} 
\end{figure*}

In this section, we describe our proposed method. 
The architecture of the model is outlined in Figure~\ref{fig:method}.
Next, we explain each stage of the pipeline and refer the reader to~\supp
for further details.

\subsection{Visual backbone}

\xpar{CNN.}
The input to the pipeline is a silent video clip of $T$ frames, $\bm{x} \in \mathbb{R}^{T \times H \times W \times 3}$.
A spatio-temporal residual CNN is applied on sub-clips of $5$ frames (i.e. 0.2s) with a unit frame stride,
to extract visual spatial feature maps $\bm{f} \in \mathbb{R}^{T\times h \times w \times c }$.
For our best model, $H=W=96$, $(h, w) = (H/4, W/4) = (24, 24)$, and $c=128$.

\xpar{Visual Transformer Pooling (VTP).}
The CNN feature map $\bm{f}_t \in \mathbb{R}^{  hw \times c } $ corresponding to every input frame $t\in \{1,\dots,T\}$ is processed individually by a shared Visual Transformer Pooling (VTP) block.
The feature map is first flattened to $\bm{f}_t \in \mathbb{R}^{  h \times w \times c } $ and projected to a desired Transformer feature dimension $d$ to get $\bm{f}_t \in \mathbb{R}^{  hw \times d } $. Then,
spatial positional encodings (SPE) are added to it;
the result is passed through an encoder consisting of $N_{VTP}$ Transformer
layers, to get an enhanced self-attended feature map
$$
\bm{z}_t = encoder_{v}( \bm{f}_t + SPE_{1:hw}) \in \mathbb{R}^{hw \times d}.
$$
A learnable query vector $\bm{Q_{att}} \in \mathbb{R}^{d \times 1 } $ is then used to
extract a visual attention mask
$$
\bm{a}_t = softmax ( \bm{Q_{att}}^\top \ \bm{z}_t ) \in \mathbb{R}^{ hw \times 1 }.
$$

The attention mask is used to compute a weighted average over the self-attended feature map 
$$
\bm{g}_t = \frac{1}{hw} \sum_{u = 1}^{hw} a_t^u z_t^u \in \mathbb{R}^{d}
$$
where $a^u_t$ and $z^u_t$ denote the feature and attention weight respectively, associated with frame $t$ and
location $u\in \{1,\dots,hw\}$.
By stacking the resulting vectors $\bm{g}_t$ in time,
we obtain an embedding sequence $\bm{g} = ( \bm{g}_1, \bm{g}_2 \cdots, \bm{g}_T ) \in \mathbb{R}^{ T \times d} $ 
which contains a compact spatio-temporal representation for every input frame.

\xpar{Transformer encoder-decoder.}
An encoder-decoder Transformer model is used to predict a text token sequence $s = (s_1, s_2 \cdots, s_{T_{dec}})$
from the source video embedding sequence  $\bm{g}$,
one token at a time:
temporal positional encodings (PE) are added to $\bm{g}$, and the result is input 
to an encoder, which consists of $N_{enc}$ multi-head Transformer layers, to produce a self-attended embedding sequence 
$$
\bm{g_{enc}} =
\textsc{encoder}(\bm{g} + PE_{1:T}) \in \mathbb{R}^{T \times d}.
$$
The decoder, which consists of $N_{dec}$ Transformer layers,
then attends on this sequence and predicts the output text token sequence
$s$ in an auto-regressive manner, by factorising its joint probability:
\begin{align}
  \log p(s | \bm{x}) =  \sum_{t=1}^{T_{dec}} \log p(s_t  | \ {\bm{g_{enc}}(\bm{x}), s_{1:t-1}} ) 
\end{align}
where positional encodings have also been added to the auto-regressive decoder inputs
as in \cite{Vaswani2017}.

The text sentences are encoded into token sequences (and vice versa tokens are decoded into text) using a sub-word level tokeniser, in particular
WordPiece~\cite{Wu16}. We tried other sub-word tokenizations such as Byte-Pair-Encoding (BPE) that is used in GPT2~\cite{radford2019language}, but it performed worse compared to using WordPiece.

\xpar{Beam search decoding and rescoring.}
Decoding is performed with a left-to-right beam search of width $B$. We also decode a second time after flipping all the input video frames horizontally.
Additional language knowledge can be incorporated by using an external language model (LM)
to re-score~\cite{Chan16} the $2 \times B$-best hypotheses $S = \{ s_{1} \cdots s_{B}; s^{h}_{1} \cdots s^{h}_{B}\}$ that the beam searches result
in, and obtain the highest scoring one as the final sentence prediction:
$$
s_{best} = \argmax_{s \in S} 
\left[ \ \alpha \log p(s | \bm{x}) + (1-\alpha) \log p_{LM}(s) \ \right]
$$

Here, $s^{h}_{1} \cdots s^{h}_{B}$ denotes the beam sequences after horizontally flipping the input. We found that additional test-time augmentations such as small degrees of rotation and/or color jitters did not improve the results.

\subsection{Training} \label{sec:training}

\xpar{Optimisation objective.}
Given a training dataset $\mathcal{D}$ consisting of pairs $(x, s^*)$ of video clips and their
ground truth transcriptions,
the model is trained to maximise the log likelihoods of the transcriptions by optimising the following objective
\begin{align}
  \mathcal{L} = - \mathbb{E}_{(x, s^*) \in \mathcal{D}}  \log p(s^*|\bm{x})
  \label{eqn:mil-nce}
\end{align}

\xpar{Teacher forcing.}
To accelerate training, we follow common practice for sequence-to-sequence training with Transformers,
and feed in the previous ground truth token as the decoder input at every step, instead of using auto-regression.
The tokens are fed into the decoder via a learnable embedding layer.

\xpar{Training protocol.}
Training is performed in two stages. 
First the whole network is trained end-to-end on short phrases of $2$ words.
Following~\cite{Chung16,Afouras19}, we use frame word-boundaries to crop out training
samples from all the possible combinations of $2$ consecutive words in the dataset, which provides natural augmentation.
Upon convergence, we freeze the visual backbone, then pre-extract and dump the visual features
for all the samples.
In the second training stage that follows, we train the encoder-decoder sub-network 
on all possible sub-sequences (n-grams) of length $2$ or larger that can be generated by combining consecutive
word utterances in the dataset.

\xpar{Discussion.}
We note that our training protocol is much simpler than the ones commonly used in prior works~\cite{Chung16,Afouras19,petridis2018audio}, since
(i) the same network and loss are used during the backbone pre-training stage, which provides a good initialization of the entire network and enables a smooth transfer; this is in contrast to other works that pre-train with a different proxy loss and require a separate word
classification head which is subsequently discarded; 
and (ii), the second stage is significantly simpler to implement
and requires a single run, unlike curriculums that gradually increase the length of the
training sentences and usually require a complicated tuning process with multiple manual restarts to achieve the best results. We observed that our proposed second stage training setup matches the performance of the complicated curriculum strategy used in previous works, while being more efficient in terms of training time and manual efforts. 

\begin{table*}
\setlength{\tabcolsep}{6.0pt}
\begin{center}
\begin{tabular}{ l l r | r r } 
 \toprule
 & \multicolumn{2}{c}{\textbf{Training}}  &  \multicolumn{2}{c}{\textbf{Evaluation}} \\

\textbf{Method}                                 & Datasets used  & Total \# hours  & LRS2   & LRS3  \\  
 \midrule                                                                                                               
 LIBS \cite{zhao2019hearing}                    & LRS2, LRS3              & 698 & 65.3  &  -    \\  
 Hyb. CTC/Att. \cite{petridis2018audio}         & LRS2, LRW                       & 389  & 63.5  &   -   \\ 
 TDNN \cite{Jianwei20tdnn}                      & LRS2                            & 224  & 48.9  &  -    \\  
 Conv-seq2seq \cite{zhang2019spatio} & LRS2, LRS3                                 & 698  & 51.7  &  60.1 \\  
 CTC + KD \cite{Afouras20}                      & LRS2, LRS3, VoxCeleb2$^\ddagger$& 1,032   & 51.3  &  59.8 \\  
 Hyb. + Conformer~\cite{ma21conformer}          & LRS2, LRW                       & 389  & 37.9  &  -    \\  
 Hyb. + Conformer~\cite{ma21conformer}          & LRS3, LRW                       & 639  & -     & 43.3 \\  
 Ours                                           & LRS2, LRS3                      & 698  & \textbf{28.9}     &   \textbf{40.6}   \\  
 \midrule                                                                                                       
 TM-seq2seq \cite{Afouras19}          & LRS2, LRS3, LRW, {MV-LRS}$^\dagger$ & 1,637 & \privd{48.3}  &  \privd{58.9} \\  
 CTC-V2P \cite{Shillingford18}        & {LSVSR}$^\dagger$         & 3,886 &  -    & \privd{55.1} \\  
 RNN-T \cite{makino2019recurrent}     & {YT31k}$^\dagger$         & 31,000  &  -    & \privd{33.6} \\  
 Ours                                           & LRS2, LRS3, {MV-LRS}$^\dagger$,
 TEDx\textsubscript{ext}  & 2,676 & \textbf{\privd{22.6}}  &  \textbf{\privd{30.7}}  \\  
 \bottomrule
\end{tabular}             
\end{center}
\caption{ 
Comparison of different lip reading models on the test sets of the LRS2 and LRS3 datasets in terms of Word Error Rate \% (WER, lower is better), 
along with the datasets and the aggregate number of hours used for training each model.
Our model achieves state-of-the-art results, outperforming all previous baselines when trained on
publicly available data (i.e. LRS2 and LRS3).
If we additionally use MV-LRS and TEDx\textsubscript{ext} for training, then our best model obtains results comparable with
that of~\cite{makino2019recurrent}, even though
we are only using an order of magnitude less data. This is indicative of the data efficiency of
our proposed pipeline.
$^\dagger$Large non-public labelled datasets: MV-LRS~\cite{Afouras19} contains 730 hours,
LSVSR~\cite{Shillingford18} 3.9k hours, and YT31k~\cite{makino2019recurrent} 31k hours of
transcribed video.
$^\ddagger$unlabelled dataset. Results shown in \privd{blue} have been obtained by training (partly or
entirely) on non-public data.
\label{tab:results}
}
\end{table*}

\section{Experiments}

\subsection{Data}

\xpar{LRS2 \& LRS3.}
For training and evaluation we use two publicly available sentence-level lip reading datasets: LRS2~\cite{Chung16} and LRS3~\cite{Afouras18d}.
LRS2 contains video clips from a variety of shows from British television, such as Countryfile and Top Gear; the transcribed content sums up to approximately 224 hours in total. 
LRS3 has been collected from over $5,000$ TED and TEDx talks in English, available on YouTube, totalling 475 hours. 
Both datasets have been created using a detection and tracking pipeline that produces face-cropped clips roughly centered around the speaker's talking head. All videos are available at a $224 \times 224$ pixel resolution and $25$ fps.
The datasets contain a ``pretrain'' partition that includes extensive head tracks including word boundaries that have been produced by force-aligning subtitles to the audio. 
Those word alignments enable training at any granularity.
The test sets contain only full sentences.

\xpar{Additional dataset: TEDx\textsubscript{ext}.}
In order to obtain more training data, we create a new dataset from TEDx talks downloaded from YouTube, by using a pipeline similar to~\cite{Afouras18d}.
We collect $13,211$ TEDx talks in English that are not included in LRS3. 
Unlike the videos used for the creation of LRS3 which contain manually annotated transcripts, the new videos only have the closed-captions automatically produced by the YouTube ASR system. As these captions are only approximately aligned to the audio, we use the Montreal Force Aligner~\cite{McAuliffe2017MontrealFA} to obtain accurate alignments for the word boundaries needed by our training pipeline (Section 3.3).
For the rest of the processing (face detection, tracking and cropping) we used the same pipeline as in~\cite{Afouras18d}.
The resulting training dataset contains $1,204$ hours in total over $318,459$ visual speech tracks, including text transcriptions with word boundary alignment.
We call this new training set TEDx\textsubscript{ext}. 
We note that since this pipeline does not require any manual transcriptions, the supervision comes for free, therefore it is easily scalable. However, the supervision is not as strong due to the noise in the training data introduced by the imperfect ASR transcriptions. But, as we will see, our model achieves a huge performance boost after training on this noisy data. 

\subsection{Implementation details}

During the first training stage, we apply random visual augmentations on the input frames to reduce
overfitting: the input videos are first resized to a square $160$ pixels resolution, from which a central square $96$-pixel
crop is extracted.
Random horizontal flipping and rotation (up to $10^{\circ}$) are also applied before inputting to the lip
reading pipeline.
During inference we use the central $96$-pixel crop, and only apply the horizontal flipping augmentation. 

For our best model, \ie, VTP on $(H/4, W/4)$, we set $N_{VTP}=6$ layers with 8 heads each for the encoder of the VTP module.
For computational efficiency, VTP uses the recently proposed Linear Transformer~\cite{katharopoulos2020transformers} instead of the original Transformer~\cite{Vaswani2017}. We found that this change did not lead to a drop in the recognition performance, while being much more computationally efficient. Another design choice we had to make is deciding after which CNN layer the VTP should be applied. Transformer layers are computationally expensive at higher resolution feature maps (\ie~earlier layer activations), but can capture more detailed information. Given this trade-off, we experiment with three different feature map resolutions, at $(h, w) = {(H/4, W/4), (H/8, W/8), (H/16, W/16)}$. For the latter two variants, we set the feature dimension $d = 512$. When pooling on $(h, w) = (H/4, W/4)$, we keep the compute and memory needs in check by performing two small changes: using $d=256$ for the first $3$ VTP layers, and then setting $d=512$ but down-sampling the feature map to $(H/8, W/8)$ for the remaining $3$ layers. 

The encoder-decoder Transformer contains $N_{enc}=6$ and $N_{dec}=6$ layers, each with $8$ attention heads.
We use sinusoidal positional encodings~\cite{Vaswani2017} for PE and learnable positional encodings for SPE. We use the WordPiece tokenizer of the BERT model in HuggingFace\footnote{https://huggingface.co/transformers/pretrained\_models.html}, with
a vocabulary of $30522$ tokens.
We also use an off-the-shelf pre-trained GPT2 language model for beam rescoring.
For the beam rescoring, we set hyperparameter $\alpha = 0.7$ for LRS2 and $\alpha = 0.6$ for LRS3 respectively.
We train all models with the Adam optimiser\cite{Kingma14} with $\beta_1=0.9$, $\beta_2=0.98$ and $\epsilon=10^{-9}$.
In the first stage of the training we follow a Noam learning rate schedule~\cite{Vaswani2017} for the first $50$ epochs and then reduce the learning rate by a factor of $5$ every time the validation loss plateaus, until reaching $10^{-6}$. 
For the second stage, the learning rate is initially set to $5e^{-5}$ and reduced by a factor of 5 on plateau down to $10^{-6}$.
For our best reported models on public data, the first stage of training takes approximately $14$ days on 4 Tesla v100s GPUs. The second stage takes 1.5 days on 1 Tesla v100 GPU.

\begin{figure*}
\centering                          
\includegraphics[width=\textwidth]{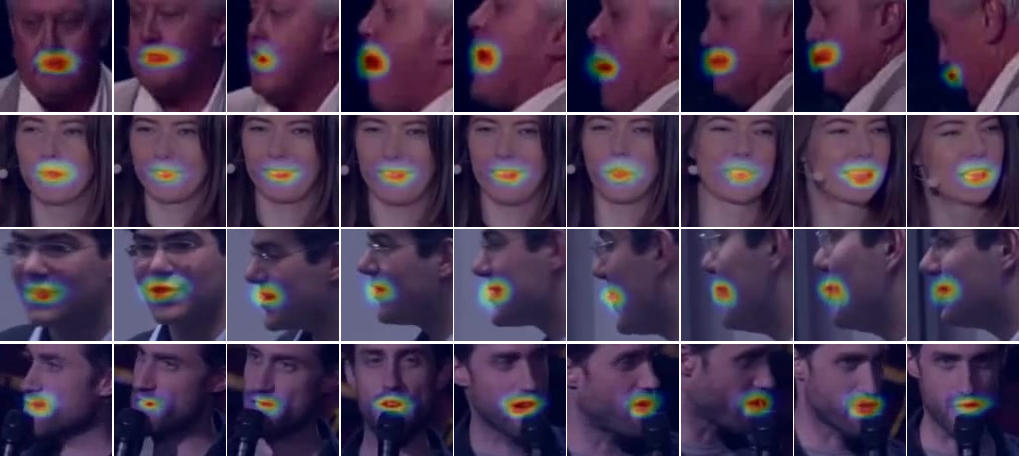} 
\caption{
  Visualization of the visual attention masks $\bm{a}$ from the VTP module superimposed on the input frames that produce them.
  The video clips used here are random samples from the LRS3 dataset.
  It is evident that the model follows the more discriminative mouth region. 
}
\label{fig:visual_maps} 
\end{figure*}

\subsection{State-of-the-art lipreading}
We compare the results of our method to existing works in~\tbl{tab:results}.
It is clear that our best model outperforms all prior work trained on public data, on both the LRS2
and LRS3 benchmarks.
In particular, compared to the strongest baseline of Ma \etal~\cite{ma21conformer} our best model
performs 9\% better on LRS2 and 2.7\% better on LRS3.
When also using MV-LRS and TEDx\textsubscript{ext} for training, we obtain a significant boost, achieving
22.6\% and 30.7\% WER for LRS2 and LRS3 respectively. We even outperform~\cite{makino2019recurrent} by a considerable margin while using $10 \times$ lesser training data. This clearly suggests that our pipeline is highly data-efficient.

\subsection{Ablations}
\xpar{Importance of each module.} We show the impact of each proposed module in the final scores, starting from
a variation of the TM-seq2seq model\cite{Afouras19}\footnote{Using the same CNN extractor as our
model for fair comparison}, and building up to our full model.
We summarize the results of this study in \tbl{tab:ablation}.
It is clear that all the proposed improvements give significant performance boosts and are largely orthogonal.
In particular, the use of WordPiece tokens contributes a $3.8\%$ absolute improvement 
on LRS2, while introducing the VTP module decreases the WER by $6.3\%$.
Using LM to rescore the beams and applying test-time horizontal flipping leads to another $1.1\%$ and $0.9\%$ improvement respectively.

\begin{table}[ht]
\begin{center}
\begin{tabular}{ l r r} 
\firsthline
\addlinespace[2pt]
\textbf{Method}                                 & WER & $\Delta$ \\ 
 \midrule                                                                                                              
   TM-seq2seq$^\dagger$ baseline                  & 41.0 & -\\  
   + WordPiece                                    & 37.2 & $-3.8$ \\ 
   + VTP                                          & 30.9 & $\mathbf{-6.3}$ \\
   + Beam LM rescoring                            & 29.8 & $-1.1$\\
   + Test-time augmentation                       &
   28.9 & $-0.9$\\
 \bottomrule
\end{tabular}             
\end{center}

\caption{ 
Ablation on the design improvements proposed in this work.
The results reported are for the test set of the LRS2 dataset.
It is clear that all the proposed components contribute
independently to the performance boost. 
$^\dagger$The baseline is an improved version of TM-seq2seq~\cite{Afouras19}
(please refer to \supp for details).}
\label{tab:ablation}
\end{table}

\xpar{VTP resolution.}
The VTP module is capable of aggregating the spatial features at arbitrary feature map resolutions. But, we show that it is more effective when operating on finer high resolution feature maps rather than coarser low resolution feature maps. This is evident in~\tbl{tab:vtp_res}, where pooling after $conv_{2,3}$ at a spatial resolution of $24\times24$ is much more effective than pooling on lower-resolution feature maps of $12\times12$ or $6\times6$.

\begin{table}[ht]
\begin{center}
\begin{tabular}{ l c r } 
\firsthline
\addlinespace[2pt]
\textbf{Method}  & \# transformer layers                               & WER  \\ 
 \midrule                                                                                                              
   Without VTP                  & 0 & 37.2\\
   VTP @ $(H/16, W/16)$ &  2           &  35.7 \\
   VTP @ $(H/8, W/8)$ & 3            & 33.8 \\ 
   VTP @ $(H/4, W/4)$ & 6                      & 30.9 \\
 \bottomrule
\end{tabular}             
\end{center}

\caption{ 
Ablation on the input spatial resolution for the VTP module. The number of Transformer layers for each stage is chosen such that the total number of parameters in the visual front-end is approximately the same. We see that pooling from higher resolution feature maps clearly leads to better results. }
\label{tab:vtp_res}
\end{table}

\xpar{Training protocol.}
Previous works~\cite{Afouras19} follow a curriculum strategy during training: the sequence length is gradually increased over the course of the training. While this protocol indeed works better, we argue that the boost in performance does not come from the curriculum learning but from something else: data augmentation. Over the course of the training process, the model gets to train on all sub-sequences (n-grams) of various lengths, and this is an effective data augmentation that reduces over-fitting. Indeed, we observe that if we simply train on all n-gram sub-sequences at once (as opposed to slow length increase), we achieve a WER of $30.92$, which is comparable to the WER of $30.91$ following a curriculum protocol. Not only does this experiment shed new light on the current understanding of the training pipeline of lip reading, it also achieves similar results while following a much simpler training process that requires significantly less manual tuning.

\subsection{Visual attention visualization}

In \fig{fig:visual_maps} we visualize the visual attention maps that the VTP module produces.
Note that the lips region is tracked very accurately while the speakers turn their heads around, even for extreme profile views. 

\section{Visual Speech Detection Application}
\begin{figure}
\centering                          
\includegraphics[width=0.6\columnwidth]{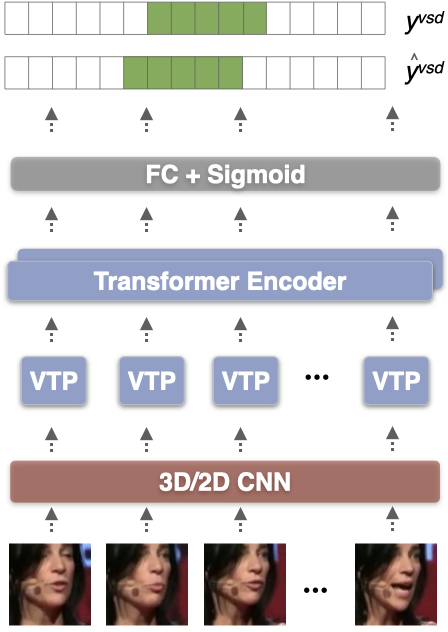} 
\caption{
Visual Speech Detection pipeline.
}
\label{fig:vsd} 
\end{figure}

We build a VSD model on top of our lip reading transformer encoder by simply adding a fully connected (FC) layer and a sigmoid activation on top of the frame-level encoder outputs to classify whether the person is speaking in that frame or not:
$$
\bm{y^{v}} =
\sigma ( \ \textsc{FC} \ (\bm{g_{enc}}) \ ) \in \mathbb{R}^{T}.
$$
The architecture is illustrated in \fig{fig:vsd}. 
We train the VSD head on top of the pre-trained lip reading encoder using a binary cross-entropy loss for every training sample:

\begin{align}
  \mathcal{L}^{v} = \frac{1}{T} \sum_{t=1}^{T} 
  y^{v}_t \log \hat{y}^{v}_t + (1-y^{v}_t) \log (1-\hat{y}^{v}_t)
  \label{eqn:vsd_loss}
\end{align}

\noindent{\textbf{Dataset and evaluation.}}
We train our VSD model on the train split of the popular AVA ActiveSpeaker dataset~\cite{Roth20ava}. This dataset is created from movies and contains $120$ videos ($2.6M$ frames) for training, $33$ videos ($768K$ frames) for validation and $109$ videos ($2M$ frames) in the test set. Each frame contains bounding box annotations for the face, along with a label indicating if the person is (i) speaking and audible, (ii) speaking but not audible, (iii) inaudible. The second class covers cases where the person is visually speaking in the background, but his voice/speech is not audible. Since we operate only on the visual frames, we combine the samples of the first two classes and train the model to perform binary classification. We initialize the weights from our pre-trained lip-reading models and fine-tune all the layers with the Adam optimizer by using a small learning rate of $10^{-6}$. 
To evaluate the performance of our model and baselines we use the mean Average Precision (mAP) metric as defined by the dataset authors~\cite{Roth20ava}; we compute the metric using the evaluation script\footnote{\url{https://github.com/activitynet/ActivityNet/blob/master/Evaluation/get_ava_active_speaker_performance.py}} that the authors supply. We also report our scores on the held-out non-public test set with the assistance of the Ava-ActiveSpeaker challenge organizers. 

\noindent{\textbf{Results.}}
We show quantitative results in~\tbl{tab:vsd}, where 
we report the VSD performance of our best VTP-based model
(corresponding to the last row of Table~\ref{tab:results}), alongside results from previous works.
It is evident that our model greatly outperforms the video-only baseline of~\cite{Roth20ava}, and even outperforms several of the recently proposed audio-visual methods (\cite{alcazar20,chung2019naver}). 
Moreover, to once again showcase the benefits of the VTP module, we also compare with an ablation of our model that uses Global Average Pooling (GAP) on top of the CNN features instead of VTP, with identical train-val-test settings. VTP outperforms this model on this task as well, by $8$ mAP points.  
We refer to \supp for qualitative video examples and examples of inference on silent videos using the whole pipeline, including VSD and lip reading.

\begin{table}[ht]
\setlength{\tabcolsep}{3.0pt}
\begin{center}
\begin{tabular}{ l c c r r} 
\firsthline
\addlinespace[2pt]
\textbf{Method}  & A & V    & mAP (val) & mAP (test) \\ 
 \midrule                                                                    
   Roth \etal~\cite{Roth20ava}       & \cmark &  \cmark     & 79.2 & 82.1 \\  
   Alcazar \etal~\cite{alcazar20}    & \cmark &  \cmark     & 87.1 & 86.7 \\  
   Chung \etal~\cite{chung2019naver} & \cmark &  \cmark     & 87.8 & 87.8 \\  
   MAAS-TAN~\cite{leon2021maas}      & \cmark &  \cmark     & 88.8 & 88.3 \\  
   TalkNet~\cite{tao2021someone}     & \cmark &  \cmark     & \textbf{92.3} & \textbf{90.8} \\  
   \midrule    
   Roth \etal~\cite{Roth20ava}       & \xmark &  \cmark     & 73.5 & 71.1 \\  
   Ours (CNN + GAP)                               & \xmark &  \cmark    & 81.4 & 80.2 \\ 
   Ours (VTP)         & \xmark &  \cmark     & \textbf{89.2} & \textbf{88.2} \\  
 \bottomrule
\end{tabular}             
\end{center}
\caption{ 
Visual Speech Detection performance on the validation (\textit{val}) and test sets of the AVA ActiveSpeaker benchmark dataset.
The \textit{A} and \textit{V} columns denote which modalities the corresponding method uses as input.
Our VTP model outperforms the baseline video-only model 
of~\cite{Roth20ava} by a large margin (over 17 mAP improvement).
In fact, we even outperform several of the recently proposed audio-visual methods (\cite{alcazar20,chung2019naver}), obtaining results close to the current state-of-the-art without using any audio.
}
\label{tab:vsd}
\end{table}

\xpar{Limitations and Ethical considerations.} We explore the limitations and failure cases of both the lip reading and the VSD models in the supplementary material, along with video examples. There we also discuss the ethical issues and the positive real-world applications of our work.

\section{Conclusion}
\label{sec:conclusion}

We have presented an
improved architecture for lip reading based on attention-based 
aggregation of visual representations as well as 
several enhancements to the training protocol, including the use of sub-word tokenisation.
Our best models
achieve state-of-the-art results,
outperforming prior work trained on public data by a significant margin,
and even industrial models trained on orders of magnitude more data.
We have also designed a Visual Speech Detection model on top of our lip reading system that obtains state-of-the-art results on this task and even outperforms several audio-visual baselines.

\balance
{\small
\bibliographystyle{ieee_fullname}
\bibliography{shortstrings,vgg_local,vgg_other,references}
}
\clearpage

\appendix
\section{Architecture Details}
In addition to the VTP implementation details in Section 4.2 of the main paper, we provide details of the CNN architecture in Table~\ref{tab:cnn_arch}. The grayed out lines denote that those were used in the CNN baseline only, and were removed when adding VTP layers after a desired feature map resolution.

\begin{table}[ht]
\setlength{\tabcolsep}{2.0pt}
  \centering
  \begin{tabular}[t]{  l r c c c r }
  \toprule
  Layer & \# Channels & Kernel & Stride & Padding  & Output dims  \\  
  \midrule
  input  & 3 &   - &  -      &   -    & $T \times 96 \times 96$  \\  
  conv\textsubscript{1,1}  & 64 & (5,5,5) & (1,2,2) & (2,2,2)  & $T \times 48 \times 48 $  \\   
  \midrule
  conv\textsubscript{2,1} & 128 & (3,3)   & (2,2) &  (1,1)  &   $T \times 24 \times 24 $  \\ 
  conv\textsubscript{2,2} & 128 & (3,3)   & (1,1) &  (1,1)  &   $T \times 24 \times 24 $  \\ 
  conv\textsubscript{2,3} & 128 & (3,3)   & (1,1) &  (1,1)  &   $T \times 24 \times 24 $  \\ 
  \midrule
  \seql{conv\textsubscript{3,1}} & \seql{256} & \seql{(3,3)}   & \seql{(2,2)} &  \seql{(1,1)}  &   \seql{$T \times 12 \times 12 $}  \\ 
  \seql{conv\textsubscript{3,2}} & \seql{256} & \seql{(3,3)}   & \seql{(1,1)} &  \seql{(1,1)}  &   \seql{$T \times 12 \times 12 $}  \\ 
  \seql{conv\textsubscript{3,3}} & \seql{256} & \seql{(3,3)}   & \seql{(1,1)} &  \seql{(1,1)}  &   \seql{$T \times 12 \times 12 $}  \\ 
  \midrule
  \seql{conv\textsubscript{4,1}} & \seql{512} & \seql{(3,3)}   & \seql{(2,2)} &  \seql{(1,1)}  &   \seql{$T \times 6 \times 6 $}  \\ 
  \seql{conv\textsubscript{4,2}} & \seql{512} & \seql{(3,3)}   & \seql{(1,1)} &  \seql{(1,1)}  &   \seql{$T \times 6 \times 6 $}  \\ 
  \seql{conv\textsubscript{4,3}} & \seql{512} & \seql{(3,3)}   & \seql{(1,1)} &  \seql{(1,1)}  &   \seql{$T \times 6 \times 6 $}  \\ 
  \midrule
  \seql{conv\textsubscript{5,1}} & \seql{512} & \seql{(3,3)}   & \seql{(2,2)} &  \seql{(1,1)}  &   \seql{$T \times 3 \times 3 $}  \\ 
  \seql{fc}                      & \seql{512} & \seql{(3, 3)}   & \seql{(1,1)} &  \seql{(0,0)}  &   \seql{$T \times 1 \times 1 $}  \\ 
  \bottomrule
  \end{tabular}

  \caption{
    Architecture details for the visual CNN backbone.
    Batch Normalization and ReLU activation are added after every convolutional layer. 
    Shortcut connections are also added at each layer, except for the first layer of every residual block -- i.e. the ones with stride $> 1$.
    The layers shown in \seql{gray}, are only used by the TM-seq2seq baseline and not in our best model.
  }
  \label{tab:cnn_arch}
\end{table}

\section{Applications and Ethical Considerations}
\label{sec:ethical}
Narrowing the gap between lip reading and ASR performance opens up opportunities for useful
applications, as stated in the introduction of the main paper: (i) improving speech recognition when the audio is corrupted in some manner; (ii) enabling silent dictation; (iii) transcribing archival silent films; (iv) helping speech-impaired individuals, e.g.\ people suffering from Lou Gehrig’s disease speak~\cite{Shillingford18}; and (v) enabling people with aphonia (loss of voice) to communicate just by using lip movements. 

But, it also raises privacy issues and the risk of potential malign uses. 
One issue that is often raised is the potential for malign surveillance, e.g.\ using CCTV
footage from public spaces to eavesdrop on
private civilian conversations.
However, this is in fact very low risk due to a number of factors:  %
we achieve a low WER on benchmarks containing video material that is professionally produced and at high resolutions and frame-rates, and under
good lighting conditions. 
Moreover the speakers are aware of being filmed and collaborate, most of the time speaking while frontally facing the camera. In contrast, 
CCTV usually operate at much lower resolution and frame rates and from unusual angles.
As shown in prior work~\cite{Chung17a,Shillingford18,makino2019recurrent}, lip reading performance
greatly deteriorates with lower frame rate or input resolutions, or when non-frontal (e.g.\ profile or overhead viewpoints) rather than
frontal speaker views are considered.

We will be making the code and pre-trained models of this work public.
This technology is already available to a small handful of corporations that have access to enough data 
and compute resources for training.
We believe that open access is important in order to accelerate progress in the field,
as well as to enable research on defences against potential adversarial attacks~\cite{ma2021detecting,gupta2021fatalread,jethanandani2020adversarial}. Finally, releasing our code and models leads to democratisation of research -- making strong models publicly available rather than to only a handful of companies can accelerate research while also making the entire process more transparent.

Overall, we believe that the benefits of the positive applications of lip reading that we have discussed (e.g.\
medical) greatly outweigh the risk of malevolent uses, the latter ones being %
inflated, 
therefore transparent research into this field should be continued and encouraged by the community.

\begin{table*}
\setlength{\tabcolsep}{6.0pt}
\begin{center}
\begin{tabular}{| m{3.8cm} | m{3.8cm} | m{3.8cm} | m{3.8cm} | m{1.5cm}} 
 \toprule
 Ground-truth Transcript & Top-1 beam prediction & Closest beam prediction & Comment \\
 \midrule
 the six highest scoring runners up will \textbf{qualify} & the six highest scoring runners will \textcolor{red}{one of them} & the six highest scoring runners will \textcolor{blue}{qualify} & Similar lip movements, language model did not disambiguate\\
 \midrule
 don't forget that \textbf{scarecrow} & don't forget that \textcolor{red}{stage} & don't forget that \textcolor{red}{stage} & Rare word \\
 \midrule
 \textbf{when there} isn't much else in the garden & \textcolor{red}{whether} it's much else in the garden & \textcolor{blue}{when there} isn't much else in the garden & \multirow{2}{=}{Lack of context leads to ranking a similar sounding incorrect phrase on top} \\
 which is relatively small \textbf{by} scottish standards & which is relatively small \textcolor{red}{but} scottish standards & which is relatively small \textcolor{blue}{by} scottish standards &\\
 \midrule
 well into november & \textcolor{red}{a} well into november & well into november & Ambiguities near the boundaries especially for very short words like `a`\\
 \midrule
 \textbf{no food} left out & \textcolor{red}{not full} left out & \textcolor{red}{not} \textcolor{blue}{food} left out & Near-homophemes\\
 \bottomrule
\end{tabular}             
\end{center}
\caption{\textbf{Error analysis:} We highlight some of the common mistakes made by our lip reading model on the unseen LRS2 test set (WER: $22.6$). We report the prediction with the best beam score, and for the purpose of the analysis, also report the beam prediction that is closest to the ground-truth but was assigned an inferior score by the model. In the last column, we describe our reasoning for the error. We can see that the model makes meaningful mistakes, and at times, it even generates a correct transcript among its beam results.      
\label{tab:lr_failures}
}
\end{table*}

\section{Lip reading error analysis}

In Table~\ref{tab:lr_failures}, we show some failure cases for our lip reading model and also make some observations for each example. We can see that the model makes meaningful errors in most of the cases. In the future, we plan to incorporate more contextual information, such as specifying possible keywords, or using a longer temporal segment. 

\section{Limitations of our Visual Speech Detection model}
We observed that the VSD model mis-classifies segments as speech when the speakers are merely expressing emotions such as crying and laughing. One such example can be seen at 2:24 - 2:25 of the supplementary video, where the speaker is crying but it is detected as speech. This is likely due to limited examples of this kind in the training data. Another limitation is that the model struggles to generalize to segment lengths beyond what is seen during training ($\le 8$ seconds). This happens due to the positional encoding, but we overcome this by extracting short overlapping clips while performing inference on a long video.

\end{document}